\documentclass{article}

\usepackage{arxiv}

\usepackage[utf8]{inputenc} 
\usepackage[T1]{fontenc}    
\usepackage{hyperref}       
\usepackage{url}            
\usepackage{booktabs}       
\usepackage{amsfonts}       
\usepackage{nicefrac}       
\usepackage{microtype}      
\usepackage{lipsum}
\usepackage{graphicx}
\usepackage{algorithm}
\usepackage{algorithmic}
\graphicspath{ {./images/} }

\title{An Approach of Adjusting the Switch Probability based on Dimension Size: A Case Study for Performance Improvement of the Flower Pollination Algorithm}

\author{
 Tahsin Aziz \\
  Department of Conputer Science and Engineering \\
  Ahsanullah University of Science and Technology \\
  Dhaka, Bangladesh \\
  \texttt{tahsinaziz.cse@aust.edu} \\
   \And
 Tashreef Muhammad \\
  Department of Computer Science and Engineering \\
  Ahsanullah University of Science and Technology \\
  Dhaka, Bangladesh \\
  \texttt{tashreef.cse@aust.edu} \\
  \And
  Md. Rashedul Karim Chowdhury \\
  \\
  B-Trac Solutions Ltd. \\
  Dhaka, Bangladesh \\
  \texttt{rkcnil@gmail.com} \\
  \And
   Mohammad Shafiul Alam \\
   Department of Computer Science and Engineering \\
   Ahsanullah University of Science and Technology \\
   Dhaka, Bangladesh \\
   \texttt{shafiul.cse@aust.edu} \\
}

\begin{document}
\maketitle
\begin{abstract}
Numerous meta-heuristic algorithms have been influenced by nature. Over the past couple of decades, their quantity has been significantly escalating. The majority of these algorithms attempt to emulate natural biological and physical phenomena. This research concentrates on the Flower Pollination algorithm, which is one of several bio-inspired algorithms. The original approach was suggested for pollen grain exploration and exploitation in confined space using a specific global pollination and local pollination strategy. As a ``swarm intelligence" meta-heuristic algorithm, its strength lies in locating the vicinity of the optimum solution rather than identifying the minimum. A modification to the original method is detailed in this work. This research found that by changing the specific value of ``switch probability" with dynamic values of different dimension sizes and functions, the outcome was mainly improved over the original flower pollination method. 
\end{abstract}


\section{Introduction}
From time to time, researchers have tried to solve different kinds of problems using various techniques to get a better result at a lower cost \cite{deb2001multi, storn1997differential, poli2007particle, aziz2019modified}. It helps to optimize the solution. Studies suggest that optimization problems are difficult to solve when computation times are considered, as in \cite{yang2010engineering, alam2011artificial}. The demand for solving problems in many sectors, such as engineering and industry, is increasing day by day. The search for the best solution is always on throughout the world. In this situation, high-difficulty problems might be solved using new and hybrid algorithms. The algorithms that are nature-inspired have made a significant impact on the optimization arena. It has influenced many real-life problems \cite{yang2010nature, alam2011recurring, alam2016differential}. When compared to certain other complex and computational segment optimization techniques, these nature-derived algorithms achieve superior results. The findings may be seen in multi-objective functions. These functions can handle NP-hard problems with numerous variables, diverse dimensions, and other challenges\cite{abbass2002pareto}. However, some optimization algorithms still have some scope for improvement. This improvement leads to optimization with higher accuracy and can be achieved for many algorithms, such as the Original Flower Pollination Algorithm (FPA)\cite{khan2012comparison}.

The process of pollinating flowers has influenced the mechanism of the FPA. Pollination, in reality, refers to the transfer of pollen grains from a flower's male anther to its female stigma\cite{alam2012diversity}. Every biological individual's purpose is to create descendants for the future. Plants are no different. Plants generate progeny in a variety of ways, including the production of seeds. In most cases, it requires a pollination process \cite{yang2013multi}.

In the original FPA, a probability factor named ``switch probability" is used. It defines the rate of global and local pollination. It is documented that when the switch probability factor is held at 0.8, the best outcome is observed \cite{walker2009flowers}. The meaning is that when the audience thinks 80\% of the pollination is local and 20\% is global, the FPA shows the best outcome \cite{sakib2014comparative}. In this paper, an observation will be made where the statements differ. It is investigated to find a better value of probability to increase the accuracy of the optimization algorithm. There has been research regarding the improvement of FPA based on the variability of the switch probability factor, \cite{ozsoydan2019analysing}. But the use of dimensions to do so is limited. Through using dimension, an improvement to the FPA algorithm will be proposed. The improved approach will change the value of the switch probability based on the value of the dimension. There is research regarding this topic of changing the value of switch-probability according to dimension \cite{wang2014flower}. But the value of it remains in the bounds of 0.6 to 0.7 in that paper.

The discussion topics in this paper are organized into different sections.

Section \ref{sec:Original Flower Pollination Algorithm} exhibits the detailed domain of the Original FPA.

Section \ref{sec: Enhanced Flower Pollination Algorithm} illustrates the proposed FPA.

Section \ref{sec:Simulation and Analysis} offers an in-depth examination of the standard functions collection. It can be seen with the algorithm's system parameters. This even assesses how well they perform.

Finally, Section \ref{sec:Conclusion} concludes its research with some guidance and suggestions for further studies.

\section{Original Flower Pollination Algorithm}
\label{sec:Original Flower Pollination Algorithm}
\subsection{Floral Pollination Features}
\label{sec: Characteristics of Flower Pollination}
Plants employ flowers as a platform to produce seeds. Pollination is the sole way for seeds to be generated. That seems to be, when pollen is transported from one bloom to another of similar species\cite{enwiki:1071070376}.

Flowers need a carrier to transport pollen. Water, insects, wind, butterflies, bats, birds, and other creatures that explore flowers are instances of carriers or vectors. ``Pollinators" are biological creatures that transport pollen from species to species\cite{yang2012flower}. Pollinators come in a wide variety of forms. It is believed that there will be around two hundred thousand different types of pollinators in the world.

Flowers often attract animals through many strategies. Animals, attracted to these techniques, usually consume them. Later on, they unintentionally conduct the process of pollination. Once the animal travels to some other flower to devour it, pollen can drop off onto the stigma, resulting in the production of offspring or pollination of the flower. There are three types of flowering plants\cite{enwiki:1070797435}. They are: \textit{hermaphrodic}, \textit{monoecious} and \textit{dioecious}.

\textit{\textbf{Hermaphrodite:}}
Hermaphrodite means these flowering plants have both sexes in the same flower.

\textit{\textbf{Monoecious:}}
Monoecious means these flowering plants have both sexes in the same plant but in different flowers.

\textit{\textbf{Dioecious:}}
Dioecious means these flowering plants are unisexual. That is male and female plants are different.

To begin the pollination procedure, pollen must be delivered from a flower's stamen to the stigma. There are two types of pollination: self-pollination and cross-pollination.

\subsubsection{Self-pollination}
Self-pollination is defined as when pollination happens inside the same plant. It occurs in two ways: autogamy and geitonogamy. 

\paragraph{Autogamy}
Here, In this process, pollen is actually transmitted to the part known as stigma of the same flower. 

\paragraph{Geitonogamy}
In this procedure, pollen is transmitted from the anther of one flower to the stigma of some other flower on the same blooming plant. 

Unless there is a mechanism to prevent it, both \textit{hermaphrodite} and \textit{monoecious} species have the ability to self-pollinate. Eighty percent of all flowering plants are hermaphroditic, whereas just five percent are monoecious. As a result, the remaining would be dioecious.

\subsubsection{Cross-pollination}
Cross-pollination is the process when pollen from one plant is transported to another.Cross pollination is a kind of pollination in which pollen from one plant's anther is transmitted to the stigma of another plant's bloom.

Pollination is categorized into abiotic pollination and biotic pollination\cite{kaur2013robust, waser1986flower}.

\textbf{Abiotic Pollination:}
Pollination by wind or propagation in water is an example of abiotic pollination\cite{kaur2012pollination}. Abiotic pollination transports pollen from one bloom to another using inanimate mechanisms like wind and water. 

Anemophily, or wind pollination, accounts for around ninety eight percent of abiotic pollination. 

A limited fraction of plants employ rain pollination which is known as Ombrophily. Heavy downpours hamper insect pollination and kill exposed blooms, but they can distribute pollen from specially suited plants.   

Hydrophily, or water pollination, needs water to transfer pollen, often as complete anthers, over the water surface to transmit dry pollen from one bloom to another.

\textbf{Biotic Pollination:}
Animals and insects transport pollen to the stigma, resulting in biotic pollination. Biotic cross-pollination may occur across large distances. Pollinators that can travel a great distance include bees, flies, birds, and bats. These pollinators, also known as vectors, are thought to be the mediators of the global pollination process\cite{colin2000varienty}. Around one hundred thousand to two hundred thousand animal species pollinate the world's two fifty thousand blooming plant species. 

Entomophily, or insect pollination, is common on plants that have evolved colorful petals and a strong aroma to attract various types of insects such as ants, flies, bees, moths, wasps, butterflies, and beetles.

Pollination is carried out by other vertebrate type animals such as birds and bats, notably sunbirds, spiderhunters, and hummingbirds, in zoophily. 

Chiropterophily, often known as bat pollination, is the process by which bats pollinate blooming plants.

Bird pollination, also known as ornithophily, is the pollination of blooming plants by birds. 

Plants evolved to utilize bats or moths as pollinators often have white petals, a strong fragrance, and blossom at night, whereas plants adapted to use birds as pollinators have red petals and generate plenty of nectar.

Flower consistency has been seen in pollinators such as butterflies and honey bees, which implies they are more likely to pass pollen to other conspecific plants. To boost their efficiency, certain flowers include unique systems for trapping pollinators.



Mammals are not typically believed to be pollinators, although bats are important pollinators, and some even specialize in this activity \cite{enwiki:1065005883}.

Xin-She Yang divides floral consistency and pollinator activity throughout the pollination process into four conditions:

\begin{enumerate}

 \item Abiotic and self-pollination are both types of local pollination.

 \item Biotic and cross-pollination are acknowledged as global pollination processes, with pollen-carrying pollinators performing L$\acute{e}$vy flights.

 \item Flower consistency can be taken into account. It is because the likelihood of reproducing is related to the resemblance of the two flowers involved.

 \item A switch probability $p$ $\in$ [0,1] governs both local and global pollination. Local pollination, in addition to physical closeness and other elements such as wind and water, can have a substantial influence on the whole pollination process.

\end{enumerate}

\subsection{Local and Global Pollination }
\label{subsub:Global Pollination and Local Pollination }
Local and global pollination procedures are the two key processes in the FPA\cite{nawjis2020hybridization} \cite{strickland2011gale}. 

Flower pollen is delivered by pollinators in the global pollination process, therefore it may go a large distances since pollinators can frequently fly and move in a greater range. This global pollination can be mathematically represented as

\begin{equation}
    x_i^{t+1}= x_i^t  + \gamma L(\lambda)(x_i^t-g^*)
\label{eq:Global Pollination}
\end{equation}

At iteration $t$ the pollen $i$ or solution vector $x_i$ is expressed by $x_i^t$. The current best solution is given by $g^*$ which has been discovered so far. To control the step size, $\lambda$ is used as a scaling factor. Hence, parameter $L(\lambda)$ is also the step size that corresponds to the strength of the pollination \cite{abdel2014novel}. Pollinators are known to travel over a long distance with variable distance steps. Here, to mimic the traveling characteristics, L$\acute{e}$vy flight should be used. Assuming $L>0$ from a L$\acute{e}$vy distribution the Equation \ref{eq:Levy Distribution} can be found.

\begin{equation}
    L \sim \frac{\lambda\Gamma(\lambda)sin(\pi\lambda/2)}{\pi}
\frac{1}{s^{1+\lambda}},\hspace{.5 cm}(S \gg S_0 > 0)
\label{eq:Levy Distribution}
\end{equation}

In Equation \ref{eq:Levy Distribution}, the standard gamma function is represented by $\Gamma(\lambda)$. For large steps $S>0$, the L$\acute{e}$vy distribution is accurate. As a result, Rules 2 and 3, which are mostly for local pollination, may be expressed as illustrated in Equation\ref{eq:Local Pollination}.

\begin{equation}
    x_i^{t+1}= x_i^t  + \epsilon(x_j^t  + x_k^t ) 
\label{eq:Local Pollination}
\end{equation}

Using Equation \ref{eq:Local Pollination}, ${x_j}^t$ and ${x_k}^t$ represent pollen from distinct blooms of the same plant species, $i$ and $j$. The formula describes flower consistency in constrained areas. If $x_j^t$ and $x_k^t$ are both from the same species or are drawn within the same population, this is equal to a local random process if a graph can be constructed $\epsilon$ from a homogenous distribution in [0,1]. 

Section \ref{subsub:Pseudo Code of Flower Pollination Algorithm} explains the pseudo code of the authors' original FPA.

\subsection{Pseudo Code of Flower Pollination Algorithm}
\label{subsub:Pseudo Code of Flower Pollination Algorithm}
In actuality, each plant may produce numerous flowers, and each flower cluster can produce millions or even billions of pollen gametes.

However, in order to reduce complexity and improve clarity, Xin-She Yang assumed that every plant has only one bloom and that each flower releases only one gamete.

As a result, distinguishing between a pollen gamete, a flower, or a plant is unnecessary.

For the sake of simplification, one pollen gamete is denoted by $x_i$. $g^*$ is the fittest solution.

The original FPA is detailed in depth in Algorithm \ref{Algorithm: Flower Pollination}.


\begin{algorithm}[htbp]
    \caption{\textbf{: Flower Pollination Algorithm}}
    \begin{algorithmic}[1]
    \STATE Objective min or max $f(x), x={(x_{1},x_{2}\ldots,x_{d})}^{t}$
    \STATE Initialize a population of n flowers/pollen gametes \\
        \quad with random solutions
    \STATE Find the best solution $g^*$ in the initial population
    \STATE Define a switch probability $p \in$ [0,1]
    \WHILE{($t <$ MaxGeneration)}	
        \FOR{$i = 1 : n$ (all n flowers in the population)}
            \IF{(rand $< p$)}
                \STATE Draw a (d-dimensional) step vector $L$ which obeys\\
                    \quad a L$\acute{e}$vy distribution
                \STATE Global pollination via $x_i^{t+1}= x_i^t + L(g^*-x_i^t )$
            \ELSE
                \STATE Draw $\epsilon$ from a uniform distribution in [0,1]
                \STATE Randomly choose $j$ and $k$ among all the solutions
                \STATE Local pollination via $x_i^{t+1}= x_i^t  +\epsilon(x_j^t - x_k^t )$
            \ENDIF
        \STATE Evaluate new solutions
        \STATE If new solutions are better,\\
            \quad update them in the population
        \ENDFOR
        \STATE Find the current best solution $g^*$
    \ENDWHILE
    \end{algorithmic}
    \label{Algorithm: Flower Pollination}
\end{algorithm}

\section{Proposed Improvement of Flower Pollination Algorithm}
\label{sec: Enhanced Flower Pollination Algorithm}
In the original algorithm, it is stated that the ``switch probability'' provides the best value at 0.8. But research has shown that by varying the values of the probability for different dimensions, a better outcome can be found.

\subsection{Real Life Situation of Global and Local Pollination}
Global pollination is more common than local pollination in reality. It helps with gene diversity. The local pollination system is only involved where pollinators are not reliable \cite{enwiki:1065005883} or where local pollination is more effective than global pollination. Given such facts, estimating that local pollination has a higher probability than a global one at around a 1:4 ratio is quite unnatural. It has also been observed in recent research that 0.8 does not yield the best result \cite{10.5120/ijais2019451824, 10.5120/cae2019652848}.

It is observed in mother nature that pollination occurs throughout higher dimensions, local pollination decreases, and global pollination increases significantly. Inspired by such a natural phenomenon, we tried to mimic the meta-heuristic algorithm of flower pollination and add such naturally occurring actions to it. The results came out positive and provided a satisfying result.

\subsection{Shifting Value of Switch Probability}
This paper has experimented on three different valued dimensions, 10, 30, and 50, and found that as dimension increases, the ``switch probability'' provides a better outcome if reduced. Ideally, it is tested for three values of switch probability, which are 0.1, 0.2, and 0.5, and it is found that dimension 10 works significantly better for the value of 0.5. As the dimension increased to 30, the value of 0.2 provided a better result. Finally, when dimension value 50 was checked, the best result came from using 0.1 as the switch probability.

\section{Simulation and Analysis} 
\label{sec:Simulation and Analysis}

\subsection{Benchmark Functions}
In comparison to the original flower pollination algorithm, the new technique produces superior results in a variety of circumstances. It is possible to demonstrate this using benchmark functions.

The collection of benchmark functions comprises unimodal, multimodal, high-dimensional, and low dimensional optimisation functions.

A function with only one local optimum is known as a "unimodal function."

A multimodal function is a function that has numerous local optima. In order to obtain global minima in multimodal functions, the search process must avoid being stranded in areas surrounding local minima.

This paper examines whether or not the results have improved.

Table \ref{Table:Function} shows the benchmark functions that are applied, along with their specifications.

\begin{table*}[htbp]
\tabcolsep9.0pt
\caption{Details about Benchmark functions used to perform experimental studies. For all the function $f_{min}=0.0$ is held true} 
\scriptsize
\begin{center}
\renewcommand{\arraystretch}{3.0}
\resizebox{\textwidth}{!}{\begin{tabular}{|c|c|c|c|c|c|}
\hline
\textbf{\textit{Function No}} & \textbf{\textit{Function Name}} & \textbf{\textit{Dimensionality}}& \textbf{\textit{Characteristics}}& \textbf{\textit{Search Space}}& \textbf{\textit{Function Definition}}  \\
\cline{2-6}
\hline
$f_1$  & Himmelblau       & 10, 30, 50  & \textit{Multi-modal}  & $[-5,5]^D$         & $f_5(\textit{\textbf{x}}) = \frac{1}{n}\sum_{i=0}^d i x_i^4-16x_i^2+5 x_i $      \\
\hline
$f_2$     & Griewank      & 10, 30, 50  & \textit{Multi-modal}  & $[-600, 600]^D$    & $f_1(\textit{\textbf{x}})=1+\frac{1}{4000} \sum_{i=1}^d x^2_i- \prod_{i=1}^d cos(\frac{\textit{\textbf{x}}_i}{\sqrt{i}})$ \\
\hline
$f_3$   & Step             & 10, 30, 50  & \textit{Multi-modal}  & $[-100, 100]^D$    & $f_3(\textit{\textbf{x}}) = \sum_{i=1}^d  \lfloor{x_i+0.5}\rfloor^2$     \\
\hline
$f_4$   & Sphere           & 10, 30, 50  & \textit{Uni-modal}  & $[-5.12, 5.12]^D$  & $f_2(\textit{\textbf{x}})= \sum_{i=1}^d x_i^2$   \\
\hline
$f_5$   & RosenBrock       & 10, 30, 50  & \textit{Uni-modal}  & $[-15, 15]^D$      & $f_6(\textit{\textbf{x}})=\sum_{i=1}^{d-1} ((100(x_{i+1}-x_i^2)^2+(x_i-1)^2)$ \\
\hline
$f_6$   & Zakharov         & 10, 30, 50  & \textit{Uni-modal}  & $[-5, 10]^D$       & $f_4(\textit{\textbf{x}})=\sum_{i=1}^d {x_i}^2 + (\sum_{i=1}^d 0.5ix_i)^2 + (\sum_{i=1}^d 0.5i x_i)^4$                    \\
\hline

\end{tabular}}
\label{Table:Function}
\end{center}
\end{table*}

Three uni-modal functions and three multi-modal functions are included in the benchmark set.

\subsection{Parameter Settings for the algorithms}
Both the original and updated methods are validated on the benchmark functions specified in Table \ref{Table:Function}.

The experiment was carried out in 100 separate runs.

The swarm size is fixed at 50 for both the original and modified methods.

The number of generations in dimension 10 was 1000, and the switch probability, $p$, was 0.5.

Dimension 30 had a generation number of 1500 and a $p$ value of 0.2.

The number of generations for dimension 50 was set to 2500, and the value of $p$ was set to 0.1.

This pattern was followed by all of the runs. For simulation, MATLAB R2016a was deployed.

\subsection{Experimental Results} 
Both algorithms were evaluated on six distinct benchmark functions to validate the performance of the suggested approach. The experiment was carried out in a variety of dimensions, generations, and switch probability values. Tables \ref{Table:Unimodal} and \ref{Table:Multimodal} show the best, worst, mean, median, and standard deviation values of the FPA method and the proposed improved FPA. The results were compared to see which one produced the best results. The benchmark functions were utilized for verification in this study. To show the best outcome for each dimension of each function, the boldface font is utilized.

\begin{table*}[htbp]
\tabcolsep11.5pt
\caption{Experimental analysis for multimodal functions between the original FPA and proposed FPA}
\scriptsize
\begin{center}
\renewcommand{\arraystretch}{1.5}
\resizebox{\textwidth}{!}{\begin{tabular}{|c|c|c|c|c|c|c|c|c|}
\hline
\textbf{\textit{Function Number}} & \textbf{\textit{Function Name}} & \textbf{\textit{Algorithm}} & \textbf{\textit{Dimension}} & \textbf{\textit{Best}}& \textbf{\textit{Worst}}& \textbf{\textit{Mean}}& \textbf{\textit{Median}}& \textbf{\textit{SD}} \\
\hline
\smash{\raise-25pt\hbox{1}} & \smash{\raise-25pt\hbox{Himmelblau}} & FPA & \smash{\raise-6pt\hbox{10}} & -2.59E+01 & -2.12E+01 & -2.37E+01 & -2.37E+01 & 9.49E-01\\ 
\cline{3-3}\cline{5-9}
& & Proposed FPA & & -2.60E+01 & -2.26E+01 & \textbf{-2.46E+01} & -2.47E+01 & 8.95E-01\\
\cline{3-9}
& & FPA & \smash{\raise-6pt\hbox{30}} & -3.53E+01 & -2.38E+01 & -2.87E+01 & -2.86E+01 & 2.24E+00\\ 
\cline{3-3}\cline{5-9}
& & Proposed FPA & & -3.60E+01 & -2.87E+01 & \textbf{-3.27E+01} & -3.27E+01 & 1.46E+00\\ 
\cline{3-9}
& & FPA & \smash{\raise-6pt\hbox{50}} & -4.14E+01 & -2.49E+01 & -3.30E+01 & -3.31E+01 & 2.68E+00\\ 
\cline{3-3}\cline{5-9}
& & Proposed FPA &  & -4.38E+01 & -3.62E+01 & \textbf{-3.97E+01} & -3.97E+01 & 1.37E+00\\
\hline
\smash{\raise-25pt\hbox{2}} & \smash{\raise-25pt\hbox{Griewank}} & FPA & \smash{\raise-6pt\hbox{10}} & 1.69E-01 & 4.72E-01 & 3.00E-01 & 3.05E-01 & 6.16E-02\\
\cline{3-3}\cline{5-9}
& & Proposed FPA & & 4.80E-02 & 2.11E-01 & \textbf{1.08E-01} & 1.00E-01 & 2.84E-02\\
\cline{3-9}
& & FPA & \smash{\raise-6pt\hbox{30}} & 1.66E+00 & 1.26E+01 & 5.09E+00 & 4.68E+00 & 1.95E+00\\ 
\cline{3-3}\cline{5-9}
& & Proposed FPA & & 1.14E+00 & 3.74E+00 & \textbf{1.58E+00} & 1.53E+00 & 3.20E-01\\
\cline{3-9}
& & FPA & \smash{\raise-6pt\hbox{50}} & 3.88E+00 & 2.98E+01 & 1.03E+01 & 9.82E+00 & 3.79E+00\\ 
\cline{3-3}\cline{5-9}
& & Proposed FPA &  & 1.18E+00 & 2.22E+00 & \textbf{1.54E+00} & 1.53E+00 & 1.85E-01\\
\hline
\smash{\raise-25pt\hbox{3}} & \smash{\raise-25pt\hbox{Step}} & FPA & \smash{\raise-6pt\hbox{10}} & 0.00E+00 & 1.00E+00 & 3.00E-02 & 0.00E+00 & 1.71E-01\\
\cline{3-3}\cline{5-9}
& & Proposed FPA & & 0.00E+00 & 0.00E+00 & \textbf{0.00E+00} & 0.00E+00 & 0.00E+00\\
\cline{3-9}
& & FPA & \smash{\raise-6pt\hbox{30}} & 1.24E+02 & 1.47E+03 & 5.77E+02 & 5.24E+02 & 2.74E+02\\ 
\cline{3-3}\cline{5-9}
& & Proposed FPA & & 5.40E+01 & 4.35E+02 & \textbf{1.74E+02} & 1.67E+02 & 6.93E+01\\ 
\cline{3-9}
& & FPA & \smash{\raise-6pt\hbox{50}} & 4.62E+02 & 2.76E+03 & 1.25E+03 & 1.20E+03 & 4.40E+02\\ 
\cline{3-3}\cline{5-9}
& & Proposed FPA &  & 1.16E+02 & 4.81E+02 & \textbf{2.74E+02} & 2.61E+02 & 7.26E+01\\
\hline

\end{tabular}}
\label{Table:Multimodal}
\end{center}
\end{table*}


\begin{figure}[htbp]
\centerline{\includegraphics[width=\textwidth,keepaspectratio]{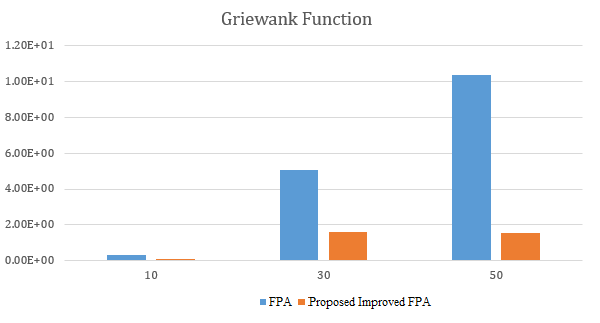}}
\caption{Experimental comparison between FPA and Proposed Improved FPA on the Griewank function for dimension 10, 30 and 50} 
	\label{fig:Griewank}
\end{figure}

\begin{table*}[htbp]
\tabcolsep11.5pt
\caption{Experimental analysis for unimodal functions between the original FPA and proposed FPA}
\scriptsize
\begin{center}
\renewcommand{\arraystretch}{1.5}
\resizebox{\textwidth}{!}{\begin{tabular}{|c|c|c|c|c|c|c|c|c|}
\hline
\textbf{\textit{Function Number}} & \textbf{\textit{Function Name}} & \textbf{\textit{Algorithm}} & \textbf{\textit{Dimension}} & \textbf{\textit{Best}}& \textbf{\textit{Worst}}& \textbf{\textit{Mean}}& \textbf{\textit{Median}}& \textbf{\textit{SD}} \\
\hline
\smash{\raise-25pt\hbox{4}} & \smash{\raise-25pt\hbox{Sphere}} & FPA & \smash{\raise-6pt\hbox{10}} & 1.72E-05 & 3.99E-04 & 1.24E-04 & 9.69E-05 & 8.35E-05\\
\cline{3-3}\cline{5-9}
& & Proposed FPA & & 3.32E-07 & 1.87E-05 & \textbf{2.86E-06} & 2.10E-06 & 3.22E-06\\
\cline{3-9}
& & FPA & \smash{\raise-6pt\hbox{30}} & 3.51E-01 & 2.75E+00 & 1.18E+00 & 1.14E+00 & 4.77E-01\\ 
\cline{3-3}\cline{5-9}
& & Proposed FPA & & 5.00E-02 & 5.33E-01 & \textbf{2.07E-01} & 1.80E-01 & 1.07E-01\\
\cline{3-9}
& & FPA & \smash{\raise-6pt\hbox{50}} & 1.20E+00 & 7.73E+00 & 2.72E+00 & 2.55E+00 & 1.06E+00\\ 
\cline{3-3}\cline{5-9}
& & Proposed FPA &  & 3.76E-02 & 2.86E-01 & \textbf{1.50E-01} & 1.46E-01 & 4.92E-02\\
\hline
\smash{\raise-25pt\hbox{5}} & \smash{\raise-25pt\hbox{Rosenbrock}} & FPA & \smash{\raise-6pt\hbox{10}} & 5.26E+00 & 2.11E+01 & 9.72E+00 & 9.29E+00 & 2.44E+00\\ 
\cline{3-3}\cline{5-9}
& & Proposed FPA & & 1.50E+00 & 9.06E+00 & \textbf{5.99E+00} & 6.42E+00 & 1.89E+00\\
\cline{3-9}
& & FPA & \smash{\raise-6pt\hbox{30}} & 5.48E+02 & 1.23E+04 & 2.75E+03 & 2.02E+03 & 2.10E+03\\ 
\cline{3-3}\cline{5-9}
& & Proposed FPA & & 8.50E+01 & 2.66E+03 & \textbf{4.35E+02} & 3.83E+02 & 3.16E+02\\ 
\cline{3-9}
& & FPA & \smash{\raise-6pt\hbox{50}} & 1.62E+03 & 2.77E+04 & 9.55E+03 & 8.44E+03 & 5.06E+03\\ 
\cline{3-3}\cline{5-9}
& & Proposed FPA &  & 1.52E+02 & 1.18E+03 & \textbf{4.71E+02} & 4.28E+02 & 1.79E+02\\
\hline
\smash{\raise-25pt\hbox{6}} & \smash{\raise-25pt\hbox{Zakharov}} & FPA & \smash{\raise-6pt\hbox{10}} & 1.21E-04 & 4.92E-03 & 1.05E-03 & 8.00E-04 & 8.64E-04\\
\cline{3-3}\cline{5-9}
& & Proposed FPA & & 1.74E-06 & 9.84E-05 & \textbf{1.97E-05} & 1.45E-05 & 1.61E-05\\
\cline{3-9}
& & FPA & \smash{\raise-6pt\hbox{30}} & 1.18E+01 & 1.02E+02 & 4.58E+01 & 4.16E+01 & 1.89E+01\\ 
\cline{3-3}\cline{5-9}
& & Proposed FPA & & 7.51E+00 & 7.04E+01 & \textbf{2.17E+01} & 2.03E+01 & 9.55E+00\\ 
\cline{3-9}
& & FPA & \smash{\raise-6pt\hbox{50}} & 5.43E+01 & 3.00E+02 & 1.51E+02 & 1.47E+02 & 5.14E+01\\ 
\cline{3-3}\cline{5-9}
& & Proposed FPA &  & 2.71E+01 & 1.20E+02 & \textbf{6.79E+01} & 6.55E+01 & 1.78E+01\\
\hline

\end{tabular}}
\label{Table:Unimodal}
\end{center}
\end{table*}


\begin{figure}[htbp]
\centerline{\includegraphics[width=\textwidth,keepaspectratio]{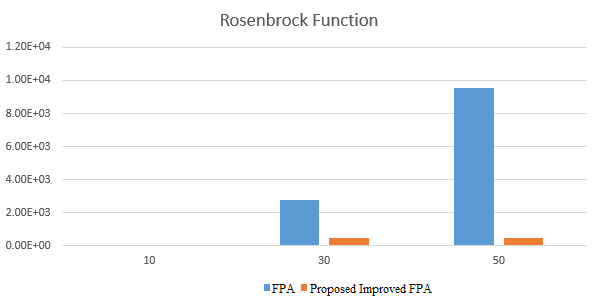}}
\caption{Experimental comparison between FPA and Proposed Improved FPA on Rosenbrock function for dimension 10, 30 and 50} 
	\label{fig:Rosenbrock}
\end{figure}

Figure $\ref{fig:Griewank}$ and Figure $\ref{fig:Rosenbrock}$  graphically show the comparison between the FPA and the proposed improved FPA in different dimensions.

From both the tables, it can be concluded that almost in all cases, the proposed improved FPA outperforms the FPA.


\section{Conclusion}
\label{sec:Conclusion}

Summarizing the experimental results on benchmark functions, it can be confidently proposed that the Flower Pollination algorithm can demonstrate better performance if the switch probability factor is kept variable instead of a fixed value. Many workspaces can be derived in the future based on this work. Such examples may include designing some technique or heuristic that automatically determines the optimum or satisfactory switch probability value based on function characteristics, modality, dimensionality, and others. The experiments conducted here are on continuous functions only, and so some future experiments may evaluate discrete functions with this proposal. Again, the results seen in this paper are based on moderately sized unimodal and multimodal functions. It provides a scope to test the proposed algorithm on large-scale and very large-scale optimization functions. Finally, the functions used in this study are theoretical. So, it would be interesting to see how the proposed algorithm performs on recent real-world and practical problems.

\bibliographystyle{unsrt}  
\bibliography{arXiv}  






\end{document}